Paper ID #196

# An ensemble Multi-Agent System for non-linear classification

Thibault Fourez[1,2*] (thibault.fourez@irit.fr),

Nicolas Verstaevel[1], Frédéric Migeon[1], Frédéric Schettini[2], Frédéric Amblard[1]

1. Institut de Recherche en Informatique de Toulouse, Université de Toulouse, CNRS, Toulouse INP, UT3, UT1, Toulouse, France

2. Citec Ingénieurs Conseil, Geneva, Switzerland

**Abstract**

Self-Adaptive Multi-Agent Systems (AMAS) transform machine learning problems into problems of local cooperation between agents. We present *smapy*, an ensemble based AMAS implementation for mobility prediction, whose agents are provided with machine learning models in addition to their cooperation rules. With a detailed methodology, we show that it is possible to use linear models for nonlinear classification on a benchmark transport mode detection dataset, if they are integrated in a cooperative multi-agent structure. The results obtained show a significant improvement of the performance of linear models in non-linear contexts thanks to the multi-agent approach.

**Keywords:**

Next-generation AI, machine learning

**1. Introduction**

Most of modern classification problems are set in a non-linear environment, i.e., in which the boundaries between classes are not hyperplanes. Because of this non-linearity, their resolution requires more complex models often called "black boxes" because of their low explicability.

In our research project, we aim to design a method to predict mobility information such as users' transport mode in real time from heterogeneous data (e.g., mobile phone data, smartphone sensors, etc.). This method must adapt quickly in a dynamic system where new transport modes and perturbations (e.g., changes in speed limits, COVID-19, etc.) may appear. Bringing up ever larger data streams requires the adoption of online learning techniques in which the model is updated with each new labeled point.

Machine learning on dynamic systems (i.e., in which the behavior of individuals, the available sensors and the classes can evolve continuously) is one of the main motivations behind the design of Multi-Agent Systems (MAS). Recent approaches propose to transform a machine learning problem into a problem of cooperation between agents in order to reduce its complexity and to allow the system to adapt to the evolutions of the individuals (Capera et al., 2003).

In this paper, we propose to use this collaborative approach to design an algorithm capable of solving supervised classification problems, some of which are non-linear, using linear classification models embedded in a multi-agent structure. The idea behind the use of linear models is to show that our contribution transforms a non-linear machine learning problem into a less complex collaborative

Integration of linear models in an ensemble multi-agent system for non-linear classification

problem (because solvable by linear models). Our algorithm must meet the following objectives:
- Guarantee good performance on nonlinear problems
- Be compatible with online learning

In the section 2, we position our approach in relation to the fields of adaptive Multi-Agent Systems (AMAS) and ensemble learning. We also present the linear models used in our experiment. We detail in section 3 the functioning of our *smapy* algorithm used in the experiment on a benchmark mobility dataset, described in section 4. Finally, we present the results obtained in the section 5 before discussing them in the section 6 and concluding.

## 2. Related work

In this section, we present the positioning of our contribution (c.f. section 3), as well as the linear models tested in this study (see section 4).

*2.1. Linear models*

Linear models are the simplest approaches in machine learning. The objective function is expressed as a linear combination of the input variables. When the target variable is qualitative, it is like looking for hyperplanes separating point clouds with different labels.

*Logistic regression.* Generalized Linear Models (GLM) (Nelder & Wedderburn, 1972) extend the domain of linear models to supervised classification problems (i.e., with a qualitative target variable) by introducing a link function in the resolution of the least squares problem. This allows to obtain probabilities of belonging to different classes. The GLM theory includes logistic regression when the link function is sigmoid.

Logistic regression is initially used to predict the probability of occurrence of an event among two modalities (binary classification). However, it is possible to generalize logistic regression to a qualitative label with $m$ modalities by performing $m$ regression successively between one class and all the others to obtain $m$ output membership probabilities.

The addition to the least squares cost function of a penalty by the $\ell_2$ norm (ridge (Hoerl & Kennard, 1970)) or $\ell_1$ norm (LASSO (Tibshirani, 1996)) of the weights vector allows to improve respectively their sparsity or the generalization of the model. These two terms are combined in the Elastic Net approach (Zou & Hastie, 2005).

*Support Vector Machines.* Support Vector Machines (SVM) (Cristianini et al., 2000) are a supervised classification technique in which the initial problem is transformed into a search for hyperplanes separating two classes via a kernel defining an intermediate space. In this new space, the problem to solve is assumed to be linear. When the kernel used is not linear, the hyperplanes of the intermediate space thus define nonlinear boundaries in the initial space, although the algorithmic resolution is that of a linear problem. However, when the kernel is linear, the resulting model is also linear.

Integration of linear models in an ensemble multi-agent system for non-linear classification

Among all possible boundaries, SVMs search for the hyperplane maximizing the confidence (i.e., the distance to the boundary) on each side. Moreover, in the case where the classes are not linearly separable in the intermediate space, it is possible to accept the separation errors for the points close to the generated frontier via a balancing parameter for the sake of generalization.

The transition from $m$-class SVMs can be done according to several strategies: *one versus all* which consists of solving $m$ binary problems (the class of interest against all others), *one versus one* which consists of training $\frac{m(m-1)}{2}$ SVMs (one per pair of classes) and then proceeding by voting for prediction, and *Crammer-Singer* strategy (Crammer & Singer, 2003) which is a reformulation of the multi-class SVM minimization problem for a quadratic computational cost.

*Passive Aggressive Algorithms.* Passive Aggressive (PA) algorithms (Crammer et al., 2006) are a family of linear online learning models that behave differently depending on whether the prediction of a new labeled point is correct or not.

In the former case, the internal model does not change (passive behavior). In the latter case, the weights are incremented so that the model predicts correctly and with a unit margin a new point identical to the last observed point (aggressive behavior). The aggressiveness of these algorithms is controlled by a coefficient in front of the weight increment at each iteration.

There are two main variants of these algorithms: PA-I when the increment is linear for the new point (Hinge cost function) and PA-II when it is quadratic (squared Hinge cost function).

*Criticism of linear models.* Linear models have the advantage of giving results that are easily interpreted by the user because the boundaries between classes are linear and the coefficients of the linear combination in the objective function give information about the importance of each input variable in the overall modeling.

However, when the system is non-linear (i.e., the point clouds of the different classes are not separable by hyperplanes), linear models perform poorly. Generating a hyperplane in a space where the classes are not linearly separable leads to misclassifications on both sides of the boundary. Algorithms such as non-linear kernel SVMs propose to transpose the initial problem to an intermediate space in which the data become linearly separable, but the coefficients associated with the variables in this space are no longer interpretable by the user.

*2.2. Ensemble learning*

Among the classification algorithms adapted to nonlinear problems, ensemble methods build classification models from a set of simple learning models, usually decision trees (CARTs). We distinguish two main types of approaches in ensemble learning: the bagging (bootstrap aggregating) and the boosting.

Integration of linear models in an ensemble multi-agent system for non-linear classification

*Bagging.* Bagging is a learning technique combining the use of bootstraps and the aggregation of prediction models. The assumption of bagging, inspired by the law of large numbers, is that averaging the predictions of several independent models reduces the variance and thus the error of the global prediction. $B$ independent models must be trained on $B$ independent data sets, which is generally not possible in practice. To overcome this problem, $B$ bootstraps are generated, i.e., $B$ samples of the same size as the initial dataset, drawn randomly (independently) and with replacement. An instance of a learning model is trained on each bootstrap, and the global model makes its predictions by averaging the intermediate models.

The Random Forest algorithm (Breiman, 2001) uses binary decision trees by adding a random draw of the input variables to be considered for each intermediate model, in order to make them even more independent. Although the geometric interpretability of the decision trees is lost, Random Forest has two importance metrics of the input variables (Mean Decrease Accuracy and Mean Decrease Gini).

*Boosting.* Unlike bagging, boosting algorithms build a model sequentially from so-called weak models. At each step, the bad points predicted by the previous model are given a higher weight when training the current model. Adaptive Boosting (AdaBoost) (Schapire et al., 1995) uses binary decision trees with a single node and a single input variable. In Gradient Boosting (Friedman, 2001), the weights of the points are no longer incremented but a cost function minimized by gradient descent allows to aggregate intermediate models to the global model. An improved version, eXtreme Gradient Boosting (XGBoost) (Chen & Guestrin, 2016), allows the user to parallelize the creation of decision trees and to optimize the prohibitive computation time of Gradient Boosting.

*Criticism of ensemble learning.* The ensemble approach allows to build a model that is efficient, not very sensitive to overfitting and generally performs better than its best internal model, especially on non-linear problems.

On the other hand, the model obtained is no longer geometrically interpretable by the user because it results from a combination of several internal models. Some algorithms, such as random forests, however, make it possible to preserve an importance score for the input variables. Moreover, ensemble methods are not suitable for online learning or dynamic systems.

### 2.3. Multi-Agent Systems

The Multi-Agent Systems (MAS) paradigm has demonstrated its ability to model and solve problems in complex, non-linear systems in which observed behaviors evolve in time and space. MAS implement autonomous entities (agents) capable of interacting with each other to solve a given problem. The key point of this theory is that each agent is generally unable to solve or even understand the problem as a whole. It is their interactions that give rise to a form of intelligence greater than the sum of their individual capacities (emergence phenomenon).

Integration of linear models in an ensemble multi-agent system for non-linear classification

*Adaptive Multi-Agent Systems.* The theory of adaptive Multi-Agent Systems (AMAS) (Capera et al., 2003) proposes a cooperative approach to interactions between agents. The design criteria presented for these interactions guarantee a satisfactory, but not necessarily optimal, result in the resolution of the problem at hand (functional adequacy theorem).

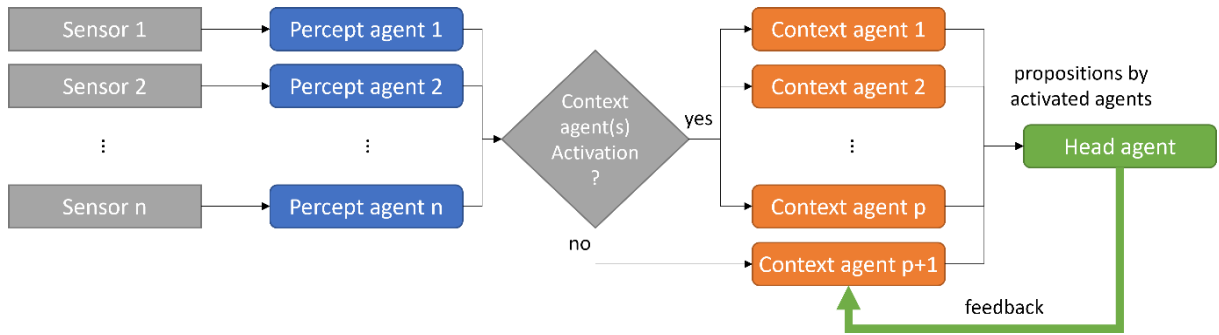

**Figure 1 - Cooperative operation of a SACL architecture in exploration phase**

*Context Learning.* Context Learning (CL) consists in exploring the space defined by the input variables of the model using cooperating agents. The AMAS for Context Learning (AMAS4CL) approach is based on the AMAS theory and more particularly on the Self-Adaptive Context Learning (SACL) (Boes et al., 2015) paradigm to define the rules of cooperation between agents and proposes a structure composed of several types of agents to explore the space of the problem variables.
Algorithms based on the SACL approach are used to solve various problems such as learning by demonstration (Verstaevel et al., 2015) or Inverse Kinematics (Dato, 2021) in robotics and optimization of the operation of a heat pump (Boes et al., 2015).
SACL architectures are typically composed of three types of agents:
- Context agents that define hypercubes of the input variable space. When a new point belongs to one of these zones, the corresponding Context agent is said to be activated and proposes a system decision according to its own knowledge
- The Percept agents which retrieve the values of the input variables (sensors) at each iteration and transmit them to the Context agents
- The Head agent which receives the proposals of the activated Context agents and sends them feedbacks

A context learning MAS has two modes of operation: exploration (learning) which consists in instantiating and arranging the Context agents thanks to the feedback from the Head agent, and exploitation (prediction) which consists in taking a decision without updating the system. The functioning of this architecture is presented in Figure 1 for the cooperative case in the exploration phase (i.e., the optimal functioning case). When the behavior of the system is not optimal with respect to the user's objectives, the situation is said to be non-cooperative (NCS). The system must adapt to maximize the cooperation between agents to return to the cooperative state. In Context Learning, this cooperation is expressed in the sizes (i.e., the dimensions of the hypercubes), positions and knowledge of the Context agents.

Integration of linear models in an ensemble multi-agent system for non-linear classification

The ELLSA architecture (Dato, 2021) is based on SACL and introduces cooperation rules to explore the space of input variables. Context agents implement linear regressions to approximate the underlying function of the problem to be solved, assumed to be locally linear on their activation zones. These internal linear regression models can also be used to solve supervised classification problems in which each Context agent is associated to a class.

*Criticism of Multi-Agent Systems.* The SACL approach has three advantages:
- The generated model is dynamic (i.e., it adapts to changes in the distribution over time of the data to be modeled) and supports online learning
- The position and the size of the agents in the space of the input variables give additional information on the studied phenomenon and a geometrical interpretation

By embedding linear regression models in the Context agents, it is possible to use the SACL architecture in supervised classification problems, but each agent is only able to predict one class. In areas of the input variable space where point clouds of different classes are overlapped, performance is severely degraded unless low-volume Context agents are generated at the risk of overfitting.

**We propose to solve this problem by taking inspiration from ELLSA's cooperation rules and providing each Context agent with an arbitrary machine learning model**, in the fashion of ensemble approaches. This allows the agents to predict several classes based on their own experience.

## 3. *Smapy*

In our contribution, we provide each Context agent with an internal machine learning model, linear or not, with the only requirement to support online learning. Each internal model is trained on the points that have activated the corresponding Context agent and thus constitutes a local modeling (in the sense of the space of input variables) of the underlying function of the problem to solve. This context learning MAS, *smapy*, has been implemented in python for the industrial needs of the research project.

### 3.1. General principle

Like other SACL type architectures, *smapy* has two modes of operation:
- The exploration during which the mapping of the space of the input variables is modified according to new available labelled observations
- The operation during which the system uses its coverage of the space of input variables to classify a new point

In both cases, the operation of the system is iterative, and each cycle starts with a new observation. During the exploration, the activated Context agents update their internal model with the last observation after they have proposed an output class to the Head agent and received feedback (positive or not). The feedback received by a Context agent allows it to update its perception of itself within the collective through a performance metric explained in the section 3.3. It also allows him to know if he has a non-cooperative behavior with respect to the objective of the system and, if necessary, to act on itself or its neighbors to return to a cooperative state (c.f. section 3.4).

Integration of linear models in an ensemble multi-agent system for non-linear classification*3.2. Agents*

In this section, we present the three types of agents involved in our SACL architecture, whose relationships have been described in Figure 1.

*Percept.* The $p$ Percept agents collect the values of the $p$ input variables of each new observation and pass them to the Context agents. They also store the observed extrema for each variable.

*Context.* A Context agent $l$ defines a hypercube in the $p$-dimensional space of input variables. For each dimension $j$, it has two parameters $r_{l,j,0}^t$ and $r_{l,j,1}^t$ that define the lower and upper bounds of an activation interval at iteration $t$. The agent can compute at any time $v_l(T)$, the volume of its activation hypercube at iteration $T$, according to the following formula: $v_l(T) = \prod_{j=1}^{p}(r_{l,j,1}^t - r_{l,j,0}^t)$

The Context agent also has a confidence level $c_l(T)$ at iteration $T$, depending on its history $\mathcal{H}_l^T$ (set of its activation cycles since its creation), its class proposals $\hat{y}_l^t$ for observations $y^t$ on this history, and two external parameters $F_+$ and $F_-$ that respectively weight the positive and negative feedbacks of the agent Head:

$$c_l(T) = \sum_{t \in \mathcal{H}_l^T}(F_+ * 1_{\hat{y}_l^t = y^t} - F_- * 1_{\hat{y}_l^t \neq y^t})$$

From its two terms, we define the score $s_l(T)$ of a Context agent at iteration $T$ using a normalization function $N_c$ which is an external parameter of *smapy*:

$$s_l(T) = N_c \circ c_l(T)$$

Finally, the Context agent has an internal classification model learned from the observations that activated it. The python implementation of *smapy* makes it possible to use models in the scikit-learn fashion if they support online learning to adapt the agent to new observations. For the rest of this paper, we define several properties of Context agents:

**Definition 1 (Expansion/Retraction).** A Context agent expands (resp. retracts) by a factor $\alpha$ when it increases (resp. decreases) its boundaries to multiply its volume by $1 + \alpha$ (resp. $1 - \alpha$).

**Definition 2 (Push).** A Context agent $l_1$ pushes a Context agent $l_2$ when $l_2$ retracts so that the previous intersection of $l_1$ and $l_2$ is completely outside $l_2$ (and thus contained only within $l_1$).

**Definition 3 (Absorption).** A context agent $l_1$ absorbs a context agent $l_2$ when $l_1$ expands to completely contain the area covered by $l_2$ and the agent $l_2$ is destroyed.

**Definition 4 (Point Exclusion).** A Context agent $l_1$ excludes an observation $y$ when $l_1$ retracts so that $y$ ends up outside $l_1$. Point exclusion is controlled by an external Boolean parameter $E$.

**Definition 5 (Overlapping Index).** The overlapping index $o_{l_1,l_2}$ is the ratio of the volume of the intersection of two agents Context $l_1$ and $l_2$ to the minimum of the volumes of these agents:

$$o_{l_1,l_2} = o_{l_2,l_1} = \frac{v_{l_1 \cap l_2}}{\min(v_{l_1}, v_{l_2})}$$

*Head.* The Head agent supervises the cooperation of the Context agents. At each iteration, it selects the class proposed by the activated Context agent with the highest score (and proceeds by vote in case of a

Integration of linear models in an ensemble multi-agent system for non-linear classification tie) and sends feedbacks to all the agents activated during the exploration phase (c.f. section 3.3). The Head agent can also create new Context agents in case of system incompetence (c.f. section 3.4).

*3.3. Feedback*

When the Context agents are activated, they propose a prediction to the Head agent. The latter selects the prediction of the agent with the highest score. During the exploration phase (learning), the Head agent sends feedbacks to the Context agents which have proposed a prediction:
- If the prediction is good (with respect to the label of the new point), then the confidence of the context agent increases by $F_+$ and it expands by a factor $\alpha$ (external parameter)
- If the prediction is bad, then the confidence of the context agent decreases by $F_-$. If point exclusion is allowed (i.e., $E$ is true), then the context agent excludes the new point. Otherwise, the Context agent's local model is fine-tuned with the new point (in the sense of online learning), and it retracts by a factor $\alpha$

*3.4. Non-cooperative situations*

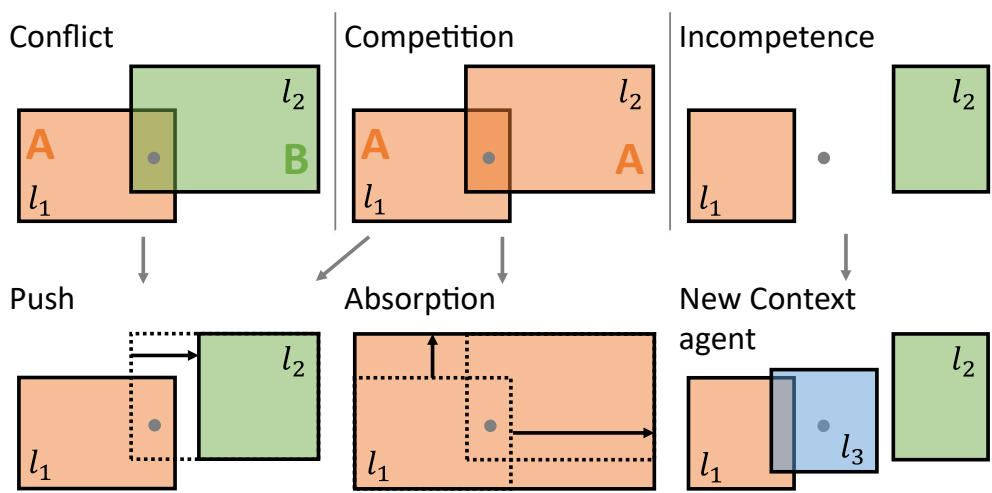

**Figure 2 - Schematic of NCS (top row) and their resolution (bottom row) for Context agents $l_1$, $l_2$ and $l_3$ predicting classes $A$ and $B$**

The objective of AMAS is to transform the initial problem into a problem of cooperation between agents. Non-cooperative situations (NCS) are states during which the behavior of the system must evolve to reach the goal set by the user. In context learning, this results in the rearrangement of the Context agents to improve the tiling of the input variable space. In this section, we present and schematize in Figure 2 the three types of NCS that can occur during the operation of *smapy* and their resolution.

*Incompetence.* Incompetence occurs when no Context agent has been activated:
- Exploration: a new Context agent is created around the new point and any NCS generated are resolved. The initial radius of the new agent is controlled by an external parameter $R$
- Exploitation: the closest Context agent to the new point (in the sense of the Euclidean distance between the point and the agent's boundary) proposes its prediction

Integration of linear models in an ensemble multi-agent system for non-linear classification

*Competition.* Competition occurs during the exploration phase when two activated Context agents propose the same prediction (in this case, the same class):
- If an overlapping threshold is defined through the external parameter $O$, and if the overlapping index of the two agents is greater than this threshold, the agent with the higher score absorbs the other
- Otherwise, the Context agent with the higher score pushes the other agent

*Conflict.* A conflict occurs during the exploration phase when two activated Context agents propose different predictions. The agent with the higher score then pushes the other agent.

## 4. Comparison of linear models alone with context learning

In this section, we present the experiment of comparing the linear models presented in section 2 (logistic regression, linear SVM, PA-I and PA-II) and instances of *smapy* with these same models inside Context agents, in a transport mode classification problem.

### 4.1. Objectives of the experiment

The motivation of this experiment is to verify if the transformation of a classification problem into a multi-agent cooperation problem without changing the machine learning techniques allows to improve the performances of the latter.

### 4.2. Input data

The experiment is conducted on the benchmark HTC transport mode detection (TMD) dataset (Yu et al., 2014) which contains smartphone sensor measurements collected during the trips of 13 users using an application developed by the authors. Five modes of transport are distinguished: Still, Walk, Run, Bike and Vehicle (including all motorized vehicles). Data cleaning and feature extraction were performed by the authors on each trip. To visualize the behavior of the Context agents in a two-dimensional space, we choose for our experiment to keep two features:
- acc std: standard deviation of the magnitude of accelerometer.
- acc FFT (peak): the index of the highest FFT value, which indicates the dominated frequency of the corresponding mode.

**Table 1 - List of value grids for the search of optimal combinations of parameters of the studied linear models (scikit-learn implementation)**

| Parameter | Grid of values | | |
|---|---|---|---|
| LOGISTIC REGRESSION & LINEAR SVM | | | |
| alpha | 0.0001 | 0.001 | 0.01 |
| penalty | $\ell_1$ | $\ell_2$ | Elastic Net |
| PA-I & PA-II | | | |
| C | 0.5 | 1.0 | 2.0 |

**Table 2 - List of value grids for finding the optimal combinations *smapy* parameters**

| Parameter | Grid of values | | |
|---|---|---|---|
| $R$ | 0.1 | 0.2 | 0.5 |
| $O$ | 0.2 | 0.5 | |
| $E$ | False | True | |
| $N_c$ | Sigmoid | | |
| $\alpha$ | 0.0 | 0.1 | 0.2 |
| $F_+$ | 1.0 | | |
| $F_-$ | 0.5 | 1.0 | 2.0 |

We obtain four instances of *smapy* and four instances of corresponding linear models, all whose parameters have been optimized by cross-validation. We then compare the optimized *smapy* instances with the linear models using two evaluation metrics:
- Classification accuracy (multi-class) averaged over the five iterations of cross-validation (step 1 for linear models, step 2 for *smapy*)
- Decision frontiers of the models (linear or *smapy*) trained with the best parameter combinations obtained by cross-validation

## 5. Results

In this section, we present comparative results between linear models alone and *smapy* instances according to the two metrics introduced previously (c.f. section 4).

**Table 3 - Comparison of the classification accuracies obtained with each model alone or in *smapy***

|  | *Model alone* | *Model + MAS* |
|---|---|---|
| *Logistic regression* | 0.65 | **0.74** |
| *Linear SVM* | 0.65 | **0.72** |
| *PA-I* | 0.58 | **0.72** |
| *PA-II* | 0.52 | **0.72** |

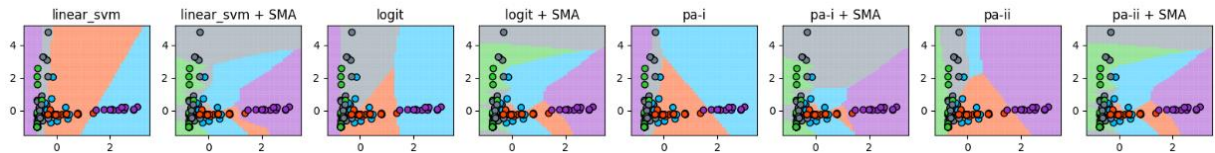

**Figure 3 - Decision boundaries for each linear model alone or in a *smapy* instance (classes are colored)**

### 5.1. Classification accuracy

The table 3 gathers the obtained classification accuracies. For each linear model tested, the MAS approach shows a significant improvement in classification accuracy.

### 5.2. Decision boundaries

We plot the decision boundaries obtained for each model (best case of cross-validations) in Figure 3. We note that the linear models alone give linear boundaries although the point clouds in the problem are not linearly separable. The MAS approach generates nonlinear boundaries that better separate the point clouds of different classes in all four cases.

## 6. Discussion

Using an MAS approach, the initial classification problem is solved locally at the Context agent level. Thus, even if the agents have internal models that can only generate linear boundaries, they have positioned and sized each other in such a way as to locally approximate a nonlinear boundary thanks to the various cooperation mechanisms presented in the section 3.

However, we can ask ourselves if the Context agents have not over-specialized locally by observing homogeneous groups of individuals (in the sense of the class). The existence of the point exclusion mechanism, although often selected by cross validation, tends to reinforce this over-specialization of agents by excluding new class points from their activation zones.

Nevertheless, the ideal behavior sought for *smapy* is to build Context agents that cover homogeneous areas of the explored input variable space, notably for reasons of explicability. There is therefore a trade-off between the geometric interpretability of the layout of the Context agents and the generalization of the system to dynamic problems in which new classes may appear.

## 7. Conclusion

Our contribution lies at the intersection of self-adaptive context learning (SACL) Multi-Agent Systems and ensemble learning methods. The proposed implementation provides each Context agent with an internal supervised classification model, as well as rules for cooperation with other agents. By choosing linear models for the Context agents, we show that it is possible to simplify a nonlinear classification problem by transforming it into a local cooperation problem within a context learning MAS. Our experimental methodology allows us to observe a significant improvement of the classification accuracy on the HTC transport mode detection dataset.

The next step is the use of *smapy* for dynamic real-world problems in which new classes may appear or disappear, with the guarantee of an interpretable prediction compared to other state-of-the-art algorithms. The hypotheses of dynamism and explicability are indeed particularly important in the field of urban planning and smart city management. The thesis project in which this paper is part of deals with mobility analysis, and experiments on automatic detection of the transport mode in an urban environment are underway.

Our main research direction on *smapy* is the possibility to use several different algorithms in the internal models of the Context agents, depending on the area of the input variables space. The idea is to exploit the strengths and weaknesses of different known algorithms to optimize prediction quality at specific locations in the space where certain models perform best. These models can be linear or non-linear, although this paper shows that internal linear models can be sufficient to obtain satisfactory results on non-linear problems.

Finally, improvements in the operating rules of *smapy* are needed to avoid over-specialization of the Context agents while maintaining the explicability and stability of the system. For this purpose, the "severity" of the NCS correction mechanisms can evolve according to the convergence of the agents' layout towards a supposedly optimal layout.

**Acknowledgement**

We thank the National Association for Research and Technology (ANRT) for the CIFRE funding of the thesis project in partnership with the *Institut de Recherche en Informatique de Toulouse* (IRIT) and the company *Citec Ingénieurs Conseil*. We also thank all the reviewers for their help and advice.

Integration of linear models in an ensemble multi-agent system for non-linear classification